\documentclass[10pt,twocolumn,letterpaper]{article}

\usepackage{cvpr}
\usepackage{times}
\usepackage{graphicx}
\usepackage{amsmath}
\usepackage{amssymb}

\usepackage{url}
\usepackage{graphicx}
\usepackage{amsmath}
\usepackage{amssymb}
\usepackage{amsthm}
\newtheorem{theorem}{Theorem}[section]
\newtheorem{lemma}[theorem]{Lemma}

\DeclareMathOperator*{\argmin}{argmin}

\DeclareMathOperator*{\sign}{sign}

\DeclareMathOperator*{\rank}{rank}
\usepackage{algorithm}
\usepackage{algorithmic}
\usepackage{subfigure}
\usepackage{url}
\usepackage{booktabs}
\usepackage{color}
\usepackage{enumitem}
\usepackage{float}

\usepackage[breaklinks=true,bookmarks=false]{hyperref}

\cvprfinalcopy 

\def\cvprPaperID{1141} 

\ifcvprfinal\fi

\begin{document}

\title{Background Subtraction via Generalized Fused Lasso Foreground Modeling}

\author{Bo Xin~~~~Yuan Tian~~~~Yizhou Wang~~~~Wen Gao\\
Nat'l Engineering Laboratory for Video Technology\\
Cooperative Medianet Innovation Center\\
Key Laboratory of Machine Perception (MoE)\\
Sch'l of EECS, Peking University, Beijing, 100871, China
}

\maketitle

\begin{abstract}
Background Subtraction (BS) is one of the key steps in video analysis. Many background models have been proposed and achieved promising performance on public data sets. However, due to challenges such as illumination change, dynamic background etc. the resulted foreground segmentation often consists of holes as well as background noise. In this regard, we consider generalized fused lasso regularization to quest for intact structured foregrounds. Together with certain assumptions about the background, such as the low-rank assumption or the sparse-composition assumption (depending on whether pure background frames are provided), we formulate BS as a matrix decomposition problem using regularization terms for both the foreground and background matrices. Moreover, under the proposed formulation, the two generally distinctive background assumptions can be solved in a unified manner. The optimization was carried out via applying the augmented Lagrange multiplier (ALM) method in such a way that a fast parametric-flow algorithm is used for updating the foreground matrix. Experimental results on several popular BS data sets demonstrate the advantage of the proposed model compared to state-of-the-arts.
\end{abstract}

\section{Introduction}
\label{sec:intro}

Background Subtraction (BS) is often regarded as a key step in video analysis. In general, it is challenging to devise a good background model and some well-known challenges include: illumination changes, dynamic background, bootstrapping, camouflage etc. To meet these challenges, many works on BS have been proposed. In the following, we discuss some related topics. 

\noindent
\textbf{Models of BS.} From the representation perspective, most existing works could be categorized into two classes: pixel-wise modeling and frame-wise modeling. In the first category, representative methods model pixel-wise statistics of the background using mixture of Gaussian models (MoG) \cite{stauffer1999adaptive,zivkovic2006efficient,haines2012background} and neural network models \cite{culibrk2007neural,maddalena2008self} etc. Nonparametric models are also proposed for improved efficiency \cite{barnich2011vibe}. The pixel-wise models are prone to resulting in fragmentary foregrounds, i.e. there are both ``holes" in the foregrounds and false positive pixels from the background. 
Whereas, the models of the second category, i.e. frame-wise models, usually achieve better performance by exploring structure information of the background. These works can be generally viewed as follow-ups of the celebrated eigen-background model proposed in \cite{oliver2000bayesian}, of which
the key assumption is that when camera motion is small, the matrix consists of background frames is approximately low-rank \cite{oliver2000bayesian,candes2011robust}. Hence, these models project video frames onto the subspace spanned by the eigen-vectors associated with the largest eigen-values of the matrix composed of all the frames of a video sequence. The recovered signal in the subspace is regarded as background and the residual is assumed to be foreground. Although structure information acquired by such holistic models helps to improve the integrity of the recovered background, such improvement can be limited in many situations due to the neglect of foreground structural prior at the same time. Thus deliberate post-processing steps are often needed e.g. in \cite{brutzer2011evaluation,haines2012background,nurhadiyatna2013background}. \textit{Therefore, is there a way to quest for intact structured foregrounds, which in turn can benefit background estimation?}

\noindent
\textbf{Learning in BS.} In some scenarios, when pure background frames are available, learning background models can be achieved in a supervised manner. We call this situation the {\em supervised model learning (SML)} case. Many pixel-wise models belong to this category and they learn/update the models from given background pixels. Whereas frame-wise models, due to their blind decomposition origin of the eigen-background, tend to neglect this piece of information. 

In many other situations, foreground background coexist in each frame. We call this situation the {\em unsupervised model learning (UML)} case and it is more challenging than the SML case. In practice, however, pixel-wise models are learnt based on the frames ahead of the test frame, regardless of such an existence of foregrounds. Consequently, in the case of UML, the pixel-wise models are less robust compared to the frame-wise counterpart, since the latter can exploit the holistic structure prior of background frames without knowing explicit background labels.

Considering the two learning cases, \textit{is there a framework that unifies both SML and UML situations of BS?}

\noindent
\textbf{The Proposed Model.} To address the above two questions, we devise a BS model that explicitly models the cohesion structure of the foregrounds in addition to the background structural prior, and propose a unified framework that solves both UML and SML cases of the BS  problem.

Notice that the foregrounds in a video sequence often correspond to meaningful objects such as people, cars, etc., therefore, the foreground pixels are usually both spatially connected as well as sparse if their sizes are relatively small w.r.t. the background scene. We realize such generic foreground structural priors by adopting an adaptive version of the generalized fused lasso (GFL) regularization in \cite{bo2014gfused}.
GFL can be viewed as a combination of the $l_1$ norm of both the variable values and their pairwise differences, i.e. the total variation (TV) penalty \cite{jordan1881serie}. By further modeling the connection/fusion strength between pixel pairs according to their similarity, (which is a strong prior in semantic segmentation \cite{boykov2006graph}), our foreground model can be considered as a flexible structural prior model without any pre-defined organization of the pixels. Specifically, we denote each frame as a vector and the sequence of frames as a matrix concatenating all the frame vectors. We assume that the observed matrix is a summation of a background matrix and a foreground matrix. Thus, by inducing a low-rank term of the background matrix and the GFL term for all foreground vectors, we formulate BS as a matrix decomposition problem. In this way, the proposed model exploits structure information from both the background and the foreground.

To harness the availability of pure background frames in the SML situation, we derive a special case of the proposed formulation. This is done by explicitly adding constraints such that part of the observed matrix equals to the given background matrix. We further assume that the unknown background vectors of the testing frames lie in the span of the given background matrix, which is itself a low-rank matrix. In this way, we show that the resulted optimization is equivalent to a sparse estimation problem.

From the perspective of optimization, the derived objective and constraints form a new problem. 
We propose an iterative algorithm by applying the augmented Lagrange multiplier (ALM) method, which alternatively updates either the background matrix or the foreground matrix. When updating the background, singular value thresholding (SVT) \cite{cai2010singular} is applied for UML and fast iterative softhreshing (FISTA) \cite{beck2009fast} is applied for SML. While updating the foreground, we solve the fused optimization with a fast parametric-flow algorithm \cite{gallo1989fast}. The idea behind this alternation is that, simultaneous estimation of the foreground and the background can reinforce each other. Indeed, experiments show that the proposed model achieves better than state-of-the-art performance on several popular data sets including both natural and synthetic videos.

\noindent
\textbf{Related Works.} In \cite{candes2011robust}, the robust principle component analysis (RPCA) model was applied to solve the BS problem. From the standpoint of BS per se, RPCA can be viewed as an extension of the eigen-background model where explicit sparse assumption of the foregrounds are taken into account, but not the connectedness. Here we introduce a stronger foreground model. In \cite{cui2012background} and \cite{xu2013gosus}, the group lasso (with overlap) regularization was applied to model the foregrounds, where the structure of foreground is assumed to be group sparse with predefined atomic group structures. These works reported improved performance over RPCA. However, in practice, experiments show that our model outperforms that of \cite{xu2013gosus}\footnote{The model in \cite{cui2012background} applied group sparsity to a trajectory representation of videos, instead of pixels we considered here. Therefore in the experiments, we focus on comparing with \cite{xu2013gosus}, which applied various of group sparsity to pixel representation and it is more recent than \cite{cui2012background}.} on all of the tested sequences. This indicates that the adaptive GFL could be a more flexible foreground structural prior compared to group lasso. In particular, Figure \ref{fig:bs_sabs2} shows such a comparison.

\noindent
\textbf{Contributions.} In summary, the contributions of this work are three folds. (1) We introduce an adaptive generalized fused lasso as a flexible structural prior to modeling foreground objects in the background subtraction problem. We show that the performance of BS can be much improved by exploiting the structure information of both the foreground and the background. (2) We propose an effective algorithm to optimize the new objective function, i.e. constrained rank minimization with GFL, by extending the method of augmented Lagrange multiplier. (3) The proposed solution to BS is a unified method which is able to solve both supervised and unsupervised learning cases depending on whether pure background frames are available, though they lead to different objectives.



\section{Proposed Background Subtraction Method}
\label{sec:meth}

\subsection{Unsupervised Model Learning}
\label{ssec:form}

We start by introducing our model for the unsupervised model learning problem, i.e. UML. Given a sequence of $n$ video frames, each frame is denoted as $\mathbf{d}^{(i)} \in \mathbb{R}^p$, $i = 1,...,n$. All data are concatenated into one matrix $\mathbf{D} \in \mathbb{R}^{p\times n}$, which is called the observation matrix.
We assume that the observation matrix is the summation of a background matrix and a foreground matrix, i.e. $\mathbf{D} = \mathbf{B} + \mathbf{F}$, where $\mathbf{B},\mathbf{F} \in \mathbb{R}^{p\times n}$ are the background matrix and the foreground matrix, respectively. Therefore, by assuming low-rank of $\mathbf{B}$ and structured sparsity of $\mathbf{F}$, we propose the following matrix decomposition objective,
\begin{equation}
\label{eq:modl0}
\begin{split}
    & \min_{\mathbf{B},\mathbf{F}} {~\rank(\mathbf{B})+ \lambda \Vert \mathbf{F} \Vert_{gfl}}\\
    & {\rm s.t.} \;\; \mathbf{D} = \mathbf{B} + \mathbf{F},
\end{split}
\end{equation}
where $\lambda \ge 0$ is a tuning parameter (controlling the relative contribution) and $\Vert \cdot \Vert_{gfl}$ is the generalized fused lasso regularization defined as
\begin{equation} \label{eq:gfl}
    \Vert \mathbf{F} \Vert_{gfl} = \sum_{k=1}^{n}\bigl\{\Vert \mathbf{f}^{(k)} \Vert_1 + \rho \sum_{(i,j)\in \mathcal{N}}{w_{ij}^{(k)}|f_i^{(k)}-f_j^{(k)}|}\bigr\},
\end{equation}
where $\mathbf{f}^{(k)}$ is the $k$th foreground vector and $\mathcal{N}$ is the spatial neighborhood set, i.e. $(i,j)\in \mathcal{N}$ when pixel $i$ and $j$ are spatially connected. Due to the $l_1$ penalties on each pixel as well as each adjacent pair of pixels, solutions of $\mathbf{f}$s tend to be both sparse and spatially connected. Here $w_{ij}$ are introduced to enhance the conventional GFL model \cite{bo2014gfused} such that $w_{ij}$ encode the strength of the fusion between neighboring pixels. In our model $w_{ij}$ is defined as
\begin{equation}
\label{eq:w}
    w_{ij}^{(k)} = \exp{\frac{-\Vert d_i^{(k)}-d_j^{(k)} \Vert_2^2}{2\sigma^2}},
\end{equation}
where $d$ is the pixel intensity. This definition of $w_{ij}$ makes it an adaptive weight encouraging spatial cohesion according to the associated pixels' intensity in the observed images. To be specific, when we observe a large difference between two neighbouring pixels, there is a high probability that this pair of pixels belongs to different segments, therefore we decrease the fusion of this pair.  $\sigma \ge 0$ is a tuning parameter empirically set. When $\sigma \rightarrow \infty$, all $w_{ij}=1$, the model reduces to the conventional GFL \cite{bo2014gfused}, where the fused term encourages pure spatial cohesion regardless of the pixel differences. When $\sigma \rightarrow 0$, all $w_{ij}=0$, the model reduces to the RPCA model \cite{candes2011robust}, where the foreground pixels are only assumed to be sparse.

For ease of optimization, the convex nuclear/trace norm is often applied to relax the matrix rank. Thus in practice, the following surrogate is considered.
\begin{equation}
\label{eq:modl}
\begin{split}
    & \min_{\mathbf{B},\mathbf{F}} {\Vert \mathbf{B} \Vert_*+ \lambda  \Vert \mathbf{F} \Vert_{gfl}}\\
    & {\rm s.t.} \;\; \mathbf{D} = \mathbf{B} + \mathbf{F},
\end{split}
\end{equation}
where $\Vert \mathbf{B}\Vert_*$ is the nuclear norm of matrix $\mathbf{B}$, i.e. the sum of the singular values of $\mathbf{B}$.

\subsection{Optimization via ALM}
\label{ssec:opt}

\begin{algorithm}[t]
   \caption{ALM algorithm for Eq. \eqref{eq:modl}.}
   \label{alg:alm}
\begin{algorithmic}[1]
   \STATE {\bfseries Input:} $\mathbf{D}\in \mathbb{R}^{p \times n}$, $\lambda>0$.
   \STATE {\bfseries Output:} $\mathbf{B}, \mathbf{F} \in \mathbb{R}^{p \times n}$.
   \STATE Initialization: Set $\mathbf{Y}_0 = \mathbf{0}$, $\mathbf{B}_0 = \mathbf{0}$, $\mathbf{F}_0 = \mathbf{0}$, $\mu_0 >0$, $\beta > 1$ and $\mu_{\max}$.
   \STATE while not converged do
    \STATE           \hspace{15mm}              $\mathbf{B}_{k+1} = \argmin_{\mathbf{B}}{ L(\mathbf{B},\mathbf{F}_k,\mathbf{Y}_k,\mu_k) } $
    \STATE           \hspace{15mm}              $\mathbf{F}_{k+1} = \argmin_{\mathbf{F}}{ L(\mathbf{B}_{k+1},\mathbf{F},\mathbf{Y}_k,\mu_k) } $
    \STATE           \hspace{15mm}              $\mathbf{Y}_{k+1} = \mathbf{Y}_k + \mu_k (\mathbf{D} - \mathbf{B}_{k+1} - \mathbf{F}_{k+1}) $
    \STATE           \hspace{15mm}             $\mu_{k+1} = \min \{\beta \mu_k, \mu_{\max}\} $
   \STATE  return $\mathbf{B}_k, \mathbf{F}_k$
\end{algorithmic}
\end{algorithm}

Eq \eqref{eq:modl} is a convex optimization problem. Off-the-shelf-solvers can be applied to solve it. However, when the dimension of $\mathbf{D}$ is large (which is often the case in BS), more efficient algorithms have to be devised. Here we employ the augmented Lagrange multiplier method \cite{bertsekas1982constrained,lin2010augmented} to solve such an equality constrained optimization. We first formulate the following augmented Lagrangian function
\begin{equation}
\label{eq:augm}
\begin{split}
    L(\mathbf{B},\mathbf{F},\mathbf{Y},\mu) = \Vert \mathbf{B} \Vert_* + \lambda \Vert \mathbf{F} \Vert_{gfl}+ \\
    \langle \mathbf{Y}, \mathbf{D} - \mathbf{B} - \mathbf{F} \rangle + \frac{\mu}{2} \Vert \mathbf{D} - \mathbf{B} - \mathbf{F} \Vert_F^2,
\end{split}
\end{equation}
where $\Vert \mathbf{\cdot} \Vert_{F}$ is the Frobenius norm, $\mathbf{Y}$ is the Lagrangian multiplier and $\mu$ is an auxiliary positive scalar. According to \cite{lin2010augmented}, the optimization problem in Eq. \eqref{eq:modl} can be solved by iteratively searching for the optimal $\mathbf{B}$, $\mathbf{F}$ and $\mathbf{Y}$ to minimize Eq. \eqref{eq:augm}. Under some rather general conditions, e.g. when $\{\mu_k\}$ is an increasing sequence and bounded, the searching process will converge Q-linearly to the optimal solution. We summarize the pseudo code in Algorithm \ref{alg:alm}\footnote{Notice that Algorithm \ref{alg:alm} is an approximate version of the original ALM. This approximation generally gives satisfactory results but converges much faster in practice \cite{lin2010augmented}.}
and discuss how to update $\mathbf{B}$ and $\mathbf{F}$ in each iteration.

\noindent
\textbf{Updating $\mathbf{B}$}. We consider the following problem
\begin{equation}
\label{eq:upB}
\begin{split}
    & \mathbf{B}_{k+1}  = \argmin_{\mathbf{B}}{ L(\mathbf{B},\mathbf{F}_k,\mathbf{Y}_k,\mu_k) } \\
                     &  = \argmin_{\mathbf{B}}{\Vert \mathbf{B} \Vert_* + \langle \mathbf{Y}_k, \mathbf{D} - \mathbf{B} - \mathbf{F}_k \rangle + \frac{\mu_k}{2} \Vert \mathbf{D} - \mathbf{B} - \mathbf{F}_k \Vert_F^2 }\\
                     & = \argmin_{\mathbf{B}} {  \frac{1}{\mu_k} \Vert \mathbf{B} \Vert_* +  \frac{1}{2} \Vert \mathbf{B} - \mathbf{M_1} \Vert_F^2 },
\end{split}
\end{equation}
where $\mathbf{M_1} = \mathbf{D} - \mathbf{F}_k + \frac{1}{\mu_k} \mathbf{Y}_k$. Eq. \eqref{eq:upB} is a standard nuclear norm minimization problem, which is known to be fast solvable via Singular Value Thresholding (SVT) \cite{cai2010singular}. According to \cite{cai2010singular}, the solution to Eq. \eqref{eq:upB} is
\begin{equation}
\label{eq:upB1}
    \mathbf{B}_{k+1} = \mathbf{U} \text{T}_{\frac{1}{\mu_k}}(\boldsymbol\Sigma) \mathbf{V}^T,~ \text{where}~ (\mathbf{U},\boldsymbol\Sigma, \mathbf{V}^T) = \text{svd}(\mathbf{M_1}).
\end{equation}
$\text{T}_{\tau}(\cdot)$ is an element-wise soft-thresholding operator, i.e. $\text{diag}(\text{T}_{\tau}(\boldsymbol\Sigma)) = [\text{t}_{\tau}(\sigma_1), \text{t}_{\tau}(\sigma_2), ... , \text{t}_{\tau}(\sigma_r) ] $ where $\text{t}_{\tau}(\sigma)$ is defined as
\begin{equation}
\label{eq:thr}
    \text{t}_{\tau}(\sigma) = \sign(\sigma)\max(|\sigma|-\tau, 0).
\end{equation}

\noindent
\textbf{Updating $\mathbf{F}$}. Now we consider the updating of $\mathbf{F}$
\begin{equation}
\label{eq:upL}
\begin{split}
   & \mathbf{F}_{k+1}   = \argmin_{\mathbf{F}}{  L(\mathbf{B}_{k+1},\mathbf{F},\mathbf{Y}_k,\mu_k) } \\
                     &  = \argmin_{\mathbf{F}}{ \lambda \Vert \mathbf{F} \Vert_{gfl} + \langle \mathbf{Y}_k, \mathbf{D} - \mathbf{B}_{k+1} - \mathbf{F} \rangle } \\
                     &~~~~~~~{+  \frac{\mu_k}{2} \Vert \mathbf{D} - \mathbf{B}_{k+1} - \mathbf{F} \Vert_F^2 }\\
                     & = \argmin_{\mathbf{F}} {  \frac{\lambda}{\mu_k} \Vert \mathbf{F} \Vert_{gfl} +  \frac{1}{2} \Vert \mathbf{F} - \mathbf{M_2} \Vert_F^2 } \\
                     & = \argmin_{\mathbf{f}^{(l)}} {\sum_{l=1}^{n} \bigl \{ { \frac{ \lambda}{\mu_k} \Vert \mathbf{f}^{(l)} \Vert_1
                    + \frac{\lambda \rho}{\mu_k}\sum_{(i,j)\in \mathcal{N}}{w_{ij}^{(l)}|f_i^{(l)}-f_j^{(l)}|}} } \\
                     &~~~~~~~{ + \frac{1}{2} \Vert \mathbf{f}^{(l)} - \mathbf{m}^{(l)} \Vert_2^2 } \bigr \},
\end{split}
\end{equation}
where $\mathbf{M_2} = \mathbf{D} - \mathbf{B}_{k+1} + \frac{1}{\mu_k} \mathbf{Y}_k$ and $\mathbf{m}^{(l)}$ is the $l$-th column of $\mathbf{M_2}$. Notice that in Eq. \eqref{eq:upL}, the optimizations of each column are independent of each other. Therefore, solving Eq. \eqref{eq:upL} equals to $n$ times of solving the following problem
\begin{equation}
\label{eq:upL1}
    \mathbf{f}^* =  \argmin_{\mathbf{f}} { \lambda_1 \Vert \mathbf{f} \Vert_1 + \lambda_2 \sum_{(i,j)\in \mathcal{N}}{w_{ij}|f_i-f_j|} + \frac{1}{2} \Vert \mathbf{f} - \mathbf{m} \Vert_2^2 },
\end{equation}
where $\lambda_1$$=$$\frac{\lambda}{\mu_k}$ and $\lambda_2$$=$$\frac{\lambda \rho}{\mu_k}$. In order to solve Eq. \eqref{eq:upL1}, according to \cite{friedman2007pathwise}, we introduce the following Lemma.
\begin{lemma}
\label{lem:soft}
    Suppose we have
\begin{equation}
\label{eq:upL2}
    \hat { \mathbf{f} }=  \argmin_{\mathbf{f}} { \lambda_2 \sum_{(i,j)\in \mathcal{N}}{w_{ij}|f_i-f_j|} + \frac{1}{2} \Vert \mathbf{f} - \mathbf{m} \Vert_2^2 },
\end{equation}
the solution to Eq. \eqref{eq:upL1}, i.e. $\mathbf{f}^*$, can be achieved by element-wise soft-thresholding such that $f^*_i = \text{t}_{\lambda_1} (\hat {f}_i)$ for $i=1,...,p$.
\end{lemma}
The proof can be shown by exploring the optimality condition of Eq. \eqref{eq:upL1} and \eqref{eq:upL2}. We provide the sketch of the proof as follows. A rigorous proof can be found in \cite{friedman2007pathwise}.
\begin{proof}
We define the objectives in Eq. \eqref{eq:upL1} and \eqref{eq:upL2} as $g_1(\mathbf{f})$ and $g_2(\mathbf{f})$ respectively. Since $\hat{\mathbf{f}}$ is the optimizer of $g_2(\mathbf{f})$, it satisfies $\partial g_2(\mathbf{f}) / \partial \mathbf{f}=0$  (sub-gradient is applied where necessary). Because the additional $\Vert \mathbf{f} \Vert_1$ term in Eq. \eqref{eq:upL1} is separable with respect to $f_i$, after applying the element-wise soft-thresholding e.g. $f^*_i = \text{t}_{\lambda_1} (\hat {f}_i)$, the resulted $\mathbf{f}^*$ satisfies $\partial g_1(\mathbf{f}) / \partial \mathbf{f}=0$.
\end{proof}
Due to Lemma \ref{lem:soft}, we can first solve Eq .\eqref{eq:upL2} and then use such an element-wise soft-thresholding technique to finally solve Eq \eqref{eq:upL1} and therefore update $\mathbf{F}$. Notice that Eq. \eqref{eq:upL2} is a continuous total variation formulation. In \cite{bo2014gfused,xin2015stable}, it is shown that Eq \eqref{eq:upL2} is equivalent to a parametric graph-cut problem which can be efficiently solved via fast flow algorithms such as the parametric-flow proposed in \cite{gallo1989fast}.

\subsection{Supervised Model Learning}
\label{ssec:form2}

In the situation where pure background frames are given (i.e. SML), we can of course still apply the same method above for background subtraction. However, by doing so, we do not fully exploit the provided information about the background. To utilized such extra information, we derive a variant of the model introduced above.

We separate the observation matrix $\mathbf{D}$ as $\mathbf{D} = [\mathbf{D}_1, \mathbf{D}_2]$, where $\mathbf{D}_1$ is the matrix of all pure background frames (the training data) and $\mathbf{D}_2$ is the matrix containing the rest frames with mixed content. The unknown $\mathbf{B}$ and $\mathbf{F}$ are separated correspondingly. We assume $\mathbf{D}_1$$ = $$\mathbf{B}_1$ and thus $\mathbf{F}_1$$ =$$ \mathbf{0}$. By applying them to Eq. \eqref{eq:modl0}, we have
\begin{equation}
\label{eq:modl1}
\begin{split}
    \min_{\mathbf{B},\mathbf{F}} & { ~\rank ([\mathbf{B}_1, \mathbf{B}_2] ) + \lambda \Vert \mathbf{F}_2 \Vert_{gfl} }\\
    & {\rm s.t.} \;\; \mathbf{D}_1=\mathbf{B}_1,~~ \mathbf{D}_2 = \mathbf{B}_2 + \mathbf{F}_2,
\end{split}
\end{equation}
We now make another assumption that $\rank(\mathbf{[B_1,B_2]}) = \rank(\mathbf{B}_1)$. The idea behind this assumption is that if we have enough pure background frames, the corresponding background vectors fully span the background subspace. 
By taking this assumption, the columns of the unknown $\mathbf{B}_2$ can be represented using linear combinations of the columns of $\mathbf{B}_1$.
Specifically, we have $\mathbf{B}_2 = \mathbf{B}_1 \mathbf{S} = \mathbf{D}_1 \mathbf{S}$, where $\mathbf{S}$ is the coefficient matrix. Thus, Eq. \eqref{eq:modl1} becomes
\begin{equation}
\label{eq:modl2}
\begin{split}
   &  \min_{\mathbf{D}_1, \mathbf{S}, \mathbf{F}_2}{~\rank ( \mathbf{D}_1[\mathbf{I}, \mathbf{S}] )+ \lambda \Vert \mathbf{F}_2 \Vert_{gfl} } \\
    & {\rm s.t.} \;\; \mathbf{D}_2 = \mathbf{D}_1 \mathbf{S} + \mathbf{F}_2.
    \end{split}
\end{equation}
Interestingly, since $\mathbf{D}_1$ is observed/given and its rank is irrelevant to the optimization. As before, we assume $\mathbf{D}_1$ to be low-rank, therefore there must exists a sparse $\mathbf{S}$. This is because each column of $\mathbf{B_2}$ can be represented as a linear combination of a small number of the columns of $\mathbf{D}_1$ (given that $\mathbf{D}_1$ is low-rank). So we can instead propose to solve
\begin{equation}
\begin{split}
\label{eq:modl3}
      &  \min_{\mathbf{S},\mathbf{F}_2} {\Vert \mathbf{S} \Vert_1+ \lambda \Vert \mathbf{F}_2 \Vert_{gfl}}\\
    & {\rm s.t.} \;\; \mathbf{D}_2 = \mathbf{D}_1 \mathbf{S} + \mathbf{F}_2,
\end{split}
\end{equation}
where $\Vert \cdot \Vert_1$ is a convex surrogate for $\Vert \cdot \Vert_0$, which counts the number of non-zero entries.

\begin{figure*}
    \centering
    \includegraphics[width = 1.6\columnwidth]{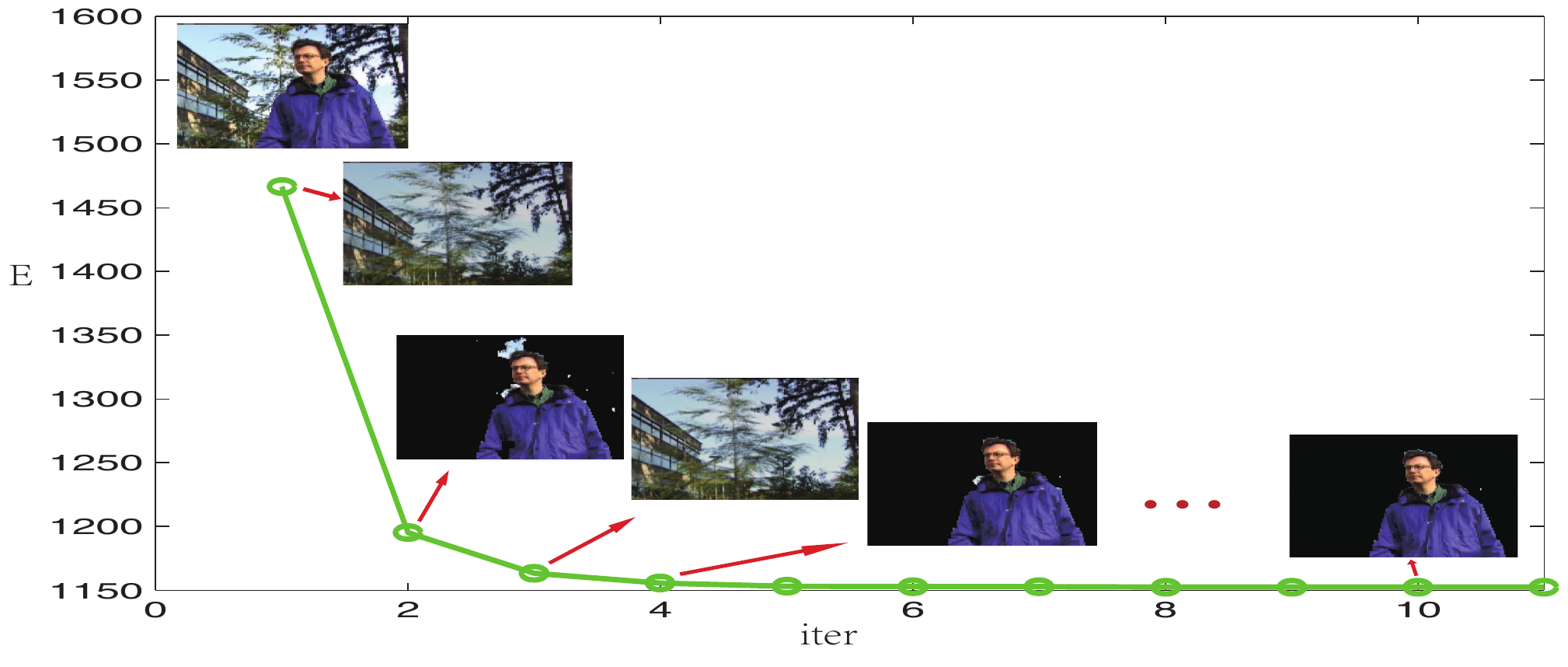}
    \caption{\small{Alternated updating of the background and the foreground. In each iteration (iter) either the background model or the foreground is updated and the objective value (the green plots) keeps decreasing until convergence.}}
\label{fig:alm}
\end{figure*}

Eq. \eqref{eq:modl3} is our SML BS model. Since it is again an equality constrained optimization, the ALM introduced above can still be applied. This time, when updating the foregrounds, the optimization is the same as before, except that we are now dealing with  $\mathbf{F}_2$ instead of $\mathbf{F}$. While updating the background, we solve the following problem
\begin{equation}
\begin{split}
\label{eq:modl4}
    \mathbf{S}_{k+1}  = \argmin_{\mathbf{S}} {  \frac{1}{\mu_k} \Vert \mathbf{S} \Vert_1 +  \frac{1}{2} \Vert \mathbf{D}_1 \mathbf{S} - \mathbf{M} \Vert_F^2 }.
\end{split}
\end{equation}

Eq. \eqref{eq:modl4} can be further decomposed into column-wise optimization and each of which is a standard Lasso \cite{tibshirani1996regression} problem. Many fast algorithms can be applied e.g. the FISTA algorithm proposed in \cite{beck2009fast}.

In summary, we can effectively solve both the UML and SML BS models by applying the ALM algorithm described in Algorithm \ref{alg:alm}. Detailed updating rules for both the background $\mathbf{B}$ and the foreground $\mathbf{F}$ are given above. Interestingly, although ALM is a general optimization method, its application to BS helps us to understand how our model alternately pursues and refines the background and the foregrounds. In Figure \ref{fig:alm}, we visualize the estimation in each iteration of ALM. We observe that the foreground estimation becomes better as the iteration goes on. This is mainly due to the synergy of the background estimation and foreground estimation. 

\section{Experiments}
\label{sec:expr}

\subsection{Data sets}

\begin{table*}[t] \small
\caption{ Brief summaries of the models compared. (Part of the descriptions are from \cite{haines2012background,brutzer2011evaluation})}
\label{tab:sum}
\begin{center}
\begin{tabular}{c|c|l}
\toprule
  methods & notation & description \\
\midrule
\cite{barnich2011vibe}   & KDE & Kernel density estimation (KDE) with a spherical kernel. Uses a stochastic history.\\
\cite{culibrk2007neural}   & G-KDE & Neural network variant of Gaussian KDE.\\
\cite{kim2004background}   & C-KDE & ’Codebook’ based; almost KDE with a cuboid kernel.\\
\cite{li2004statistical}   & Hist & Histogram based, includes co-occurrence statistics. Lots of heuristics.\\
\cite{maddalena2008self}   & Map & Uses a self organising map, passes information between pixels.\\
\cite{stauffer1999adaptive}   & MoG & Classic MoG approach. Assigns mixture components to bg/fg.\\
\cite{zivkovic2006efficient}   & R-MoG & Refinement of \cite{stauffer1999adaptive}. Has an adaptive learning rate.\\
\cite{oliver2000bayesian} & Eigen & Eigenbackground.\\
\cite{mckenna2000tracking} & Gauss & unimodal (Gaussian)\\
\cite{haines2012background} & D-MoG &  Dirichlet process Gaussian mixture Model.\\
\cite{candes2011robust} & RPCA &  Robust PCA model.\\
\cite{xu2013gosus} & G-Lasso & Online subspace learning with group lasso with overlap regularization. \\
\bottomrule
\end{tabular}
\end{center}
\end{table*}

\begin{table*}[t] \small
\caption{Results for the SABS data set, given as the F-score.}
\label{tab:accr1}
\begin{center}
\begin{tabular}{c|c|c|c|c|c|c|c|c|c|c|c}
\toprule
  & Gauss & C-KDE  & Eigen & Map & KDE & R-MoG & MoG & Hist & RPCA & G-Lasso & \textbf{Ours}\\
\midrule
F-score   & .3806& .5601 & .5891& .6672& .7177& .7232 & .7284 & .7457 &.6483 & 0.7326 &\textbf{.7775}$^*$\\
\bottomrule
\end{tabular}
\end{center}
\end{table*}

We test our model on three popular BS data sets, namely, the Wallflower\footnote{http://research.microsoft.com/en-us/um/people/jckrumm/Wallfl-ower/testimages.htm} 
data set \cite{toyama1999Wallflower} 
, the Li\footnote{{http://perception.i2r.a-star.edu.sg/bk$\_$model/bk$\_$index.html}}
data set \cite{li2004statistical} and the SABS\footnote{{http://www.vis.uni-stuttgart.de/forschung/informationsvisualisierung-und-visual-analytics/visuelle-analyse-videostroeme/sabs.html}} 
data set \cite{brutzer2011evaluation}. All together, there are 17 sequences of both natural and synthetic videos. Most well-known BS challenges are presented in these sequences, e.g. gradual/sudden illumination changes, moving background, bootstrapping, camouflage, and occlusion etc. We give a brief introduction of these data sets respectively as follows.
\begin{itemize}
    \item ``Wallflower": The Wallflower data set consists of 7 natural video sequences representing different BS challenges. The resolution of the frames is about $160 \times 120$. Manually labeled ground truth are provided. Most of the sequences have pure background frames, which can be used for SML.
    \item ``Li": The Li data set consists of 9 natural video sequences. The resolution of the frames is about $176 \times 144$. Manually labeled ground truth are provided. Part of them have pure background frames.
    \item SABS: The SABS data set is a synthetic data set and therefore provides high quality ground truth. The resolution of the frames is $800 \times 600$. Several BS challenges are synthesized to the same scene. Following \cite{haines2012background}, only the basic sequence is used. Pure background frames are available.
\end{itemize}

All three data sets are popular public data sets. Results of many existing models have been reported based on these sets. In order to evaluate the proposed model, we directly compare with the results reported in the respective papers.

\subsection{Comparison with RPCA}
\label{ssec:perf1}

Recall that the proposed model in the UML case can be reduced to RPCA when the model parameter $w_{ij} = 0$ in Eq.\ref{eq:gfl} (Section.\ref{ssec:form}). Therefore, we first show how the proposed model improves performance over the RPCA model due to GFL foreground modeling. 

\begin{figure}[h]
    \centering
    \subfigure[test image]{
    \label{fig:demo1}
        \includegraphics[width=.45\columnwidth]{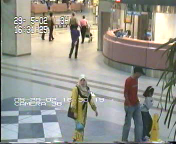}
        }   
    \subfigure[Our recovered background]{
        \includegraphics[width=.45\columnwidth]{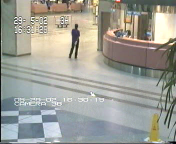}
        }
    \subfigure[Our detected foregrounds]{
        \includegraphics[width=.45\columnwidth]{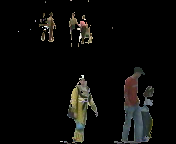}
        }  
    \subfigure[RPCA detected foreground]{
        \includegraphics[width=.45\columnwidth]{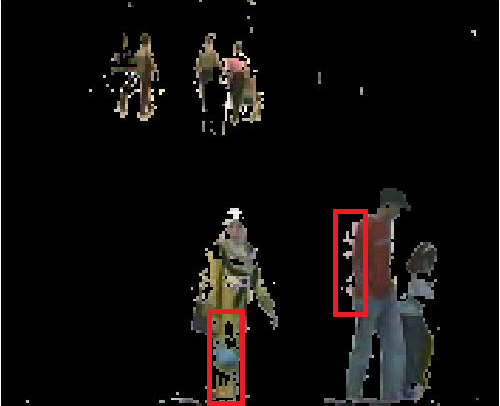}
        }
   \caption{\small{Results comparison of RCPA and our model for the airport data from the Li data set. }}
\label{fig:comrpca}
\end{figure}

Quantitative comparisons on all the sequences of the three data sets are shown in Table \ref{tab:accr1}$\sim$\ref{tab:li2}. From the comparisons we can see that the proposed model consistently outperforms RPCA.
Qualitatively, we use the same 200 frames of the airport sequence in ``Li" data set as reported in \cite{candes2011robust} to construct a head-to-head comparison, where we apply both RPCA and our model for background subtraction.  In Figure \ref{fig:comrpca}, we illustrate the BS results of the test frame used in \cite{candes2011robust}. In practice, even after fine-grid search for the best parameters, the detected foregrounds of RPCA have more ``holes" and more false positives from background than those of the proposed model. (Some obvious examples are marked by the red boxes in Figure \ref{fig:comrpca} (d). Note that these results are not post-processed.) 
Qualitative results of the whole sequence are provided on our webpage\footnote{{http://idm.pku.edu.cn/staff/wangyizhou/} }.

\begin{figure*}
	\centering
{
    \subfigure{
        \includegraphics[width=.25\columnwidth]{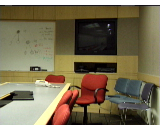}
        }    \hspace{-2.5mm}
    \subfigure{
        \includegraphics[width=.25\columnwidth]{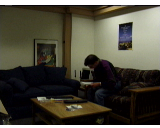}
        }    \hspace{-2.5mm}
    \subfigure{
        \includegraphics[width=.25\columnwidth]{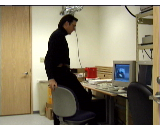}
        }    \hspace{-2.5mm}
    \subfigure{
        \includegraphics[width=.25\columnwidth]{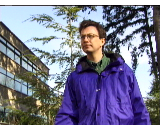}
        }    \hspace{-2.5mm}
    \subfigure{
        \includegraphics[width=.25\columnwidth]{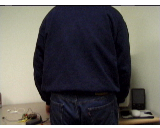}
        }    \hspace{-2.5mm}
    \subfigure{
        \includegraphics[width=.25\columnwidth]{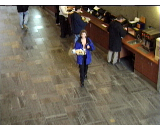}
        }    \hspace{-2.5mm}
    \subfigure{
        \includegraphics[width=.25\columnwidth]{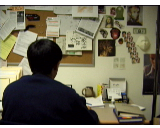}
        }    \hspace{-2.5mm}
}
    \vspace{-4mm}
{
     \subfigure{
        \includegraphics[width=.25\columnwidth]{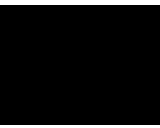}
        }    \hspace{-2.5mm}
    \subfigure{
        \includegraphics[width=.25\columnwidth]{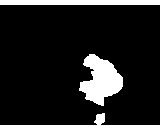}
        }    \hspace{-2.5mm}
    \subfigure{
        \includegraphics[width=.25\columnwidth]{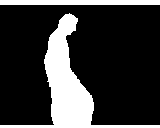}
        }    \hspace{-2.5mm}
    \subfigure{
        \includegraphics[width=.25\columnwidth]{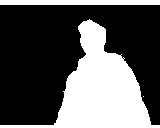}
        }    \hspace{-2.5mm}
    \subfigure{
        \includegraphics[width=.25\columnwidth]{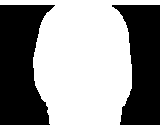}
        }    \hspace{-2.5mm}
    \subfigure{
        \includegraphics[width=.25\columnwidth]{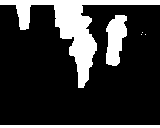}
        }    \hspace{-2.5mm}
    \subfigure{
        \includegraphics[width=.25\columnwidth]{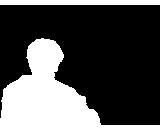}
        }    \hspace{-2.5mm}
}
    \vspace{-4mm}
    \setcounter{subfigure}{0}
{
    \subfigure[MO]{
        \includegraphics[width=.25\columnwidth]{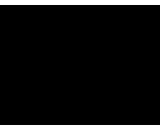}
        }    \hspace{-2.5mm}
    \subfigure[TD]{
        \includegraphics[width=.25\columnwidth]{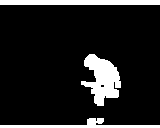}
        }    \hspace{-2.5mm}
    \subfigure[LS]{
        \includegraphics[width=.25\columnwidth]{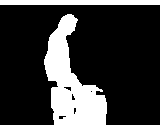}
        }    \hspace{-2.5mm}
    \subfigure[WT]{
        \includegraphics[width=.25\columnwidth]{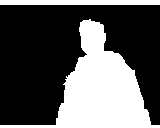}
        }    \hspace{-2.5mm}
    \subfigure[CF]{
        \includegraphics[width=.25\columnwidth]{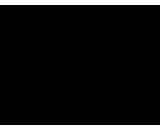}
        }    \hspace{-2.5mm}
    \subfigure[BS]{
        \includegraphics[width=.25\columnwidth]{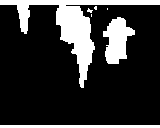}
        }    \hspace{-2.5mm}
    \subfigure[FA]{
        \includegraphics[width=.25\columnwidth]{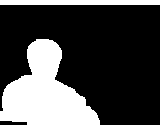}
        }
        
}
	\caption{\small{Results on Wallflower data set. From top to bottom: test images, the ground truth and the estimations of our model.}}
\label{fig_wf1}
\end{figure*}
\begin{table*}[t] \small
\caption{Results for Wallflower, given as the number of pixels that have been mix-classified.}
\label{tab:wf2}
\begin{center}
\begin{tabular}{c|c|c|c|c|c|c|c}
\toprule
 methods & MO & TD& LS & WT & CF & BS  & FA \\
\midrule
Frame Difference   & 0& 1358& 2565& 6789& 10070& 2175& 4354\\
Mean+threshold    & 0& 2593& 16232& 3285& 1832& 3236& 2818\\
Block correlation   & 1200& 1165& 3802& 3771& 6670 & 2673& 2402\\
MoG   & 0& 1028& 15802& 1664& 3496& 2091& 2972\\
Eigen   & 1065& 895& 1324& 3084& 1898& 6433& 2978\\
D-MoG              & 0& \textbf{330}& 3945& 184& \textbf{384}& 1236& 1569\\
RPCA            & 0 & 628 & 2016 & 1014  &  & 1465 & 2875 \\
G-Lasso            & 0 & 912 & 1067 & 629 &  & 1779 & 1139 \\
\textbf{Ours}          & \textbf{0}$^*$  & 418$^*$  & \textbf{686}$^*$& \textbf{166}$^*$ &  & \textbf{795}$^*$ & \textbf{192}$^*$ \\
\bottomrule
\end{tabular}
\end{center}
\end{table*}

\begin{figure*}[t]
	\centering
{
    \subfigure{
        \includegraphics[width=.2\columnwidth]{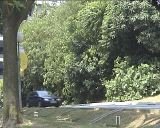}
        }    \hspace{-2.5mm}
    \subfigure{
        \includegraphics[width=.2\columnwidth]{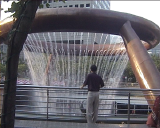}
        }    \hspace{-2.5mm}
    \subfigure{
        \includegraphics[width=.2\columnwidth]{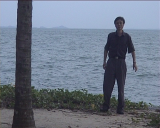}
        }    \hspace{-2.5mm}
    \subfigure{
        \includegraphics[width=.2\columnwidth]{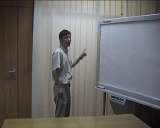}
        }    \hspace{-2.5mm}
    \subfigure{
        \includegraphics[width=.2\columnwidth]{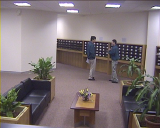}
        }    \hspace{-2.5mm}
    \subfigure{
        \includegraphics[width=.2\columnwidth]{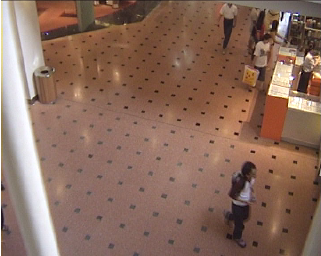}
        }    \hspace{-2.5mm}
    \subfigure{
        \includegraphics[width=.2\columnwidth]{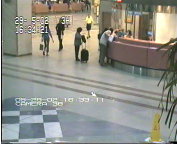}
        }    \hspace{-2.5mm}
     \subfigure{
        \includegraphics[width=.2\columnwidth]{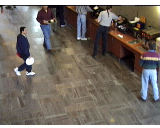}
        }    \hspace{-2.5mm}
    \subfigure{
        \includegraphics[width=.2\columnwidth]{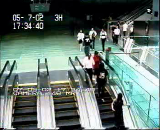}
        }   \hspace{-2.5mm}
}
         \vspace{-1mm}
{
    \subfigure{
        \includegraphics[width=.2\columnwidth]{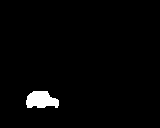}
        }    \hspace{-2.5mm}
    \subfigure{
        \includegraphics[width=.2\columnwidth]{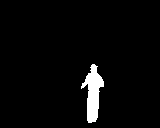}
        }    \hspace{-2.5mm}
    \subfigure{
        \includegraphics[width=.2\columnwidth]{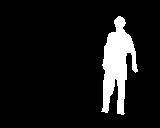}
        }    \hspace{-2.5mm}
    \subfigure{
        \includegraphics[width=.2\columnwidth]{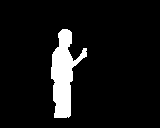}
        }    \hspace{-2.5mm}
    \subfigure{
        \includegraphics[width=.2\columnwidth]{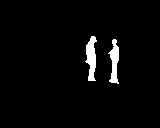}
        }    \hspace{-2.5mm}
    \subfigure{
        \includegraphics[width=.2\columnwidth]{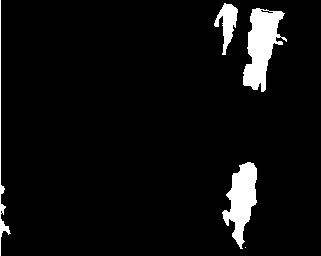}
        }    \hspace{-2.5mm}
    \subfigure{
        \includegraphics[width=.2\columnwidth]{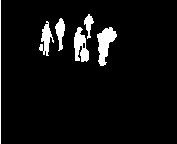}
        }    \hspace{-2.5mm}
     \subfigure{
        \includegraphics[width=.2\columnwidth]{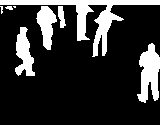}
        }    \hspace{-2.5mm}
    \subfigure{
        \includegraphics[width=.2\columnwidth]{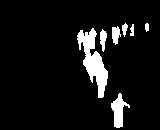}
        }   \hspace{-2.5mm}
}
    \vspace{-1mm}
    \setcounter{subfigure}{0}
{
    \subfigure[cam]{
        \includegraphics[width=.2\columnwidth]{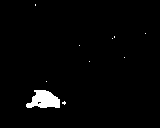}
        }    \hspace{-2.5mm}
    \subfigure[ft]{
        \includegraphics[width=.2\columnwidth]{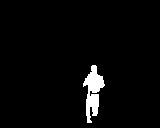}
        }    \hspace{-2.5mm}
    \subfigure[ws]{
        \includegraphics[width=.2\columnwidth]{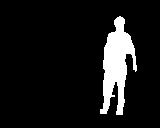}
        }    \hspace{-2.5mm}
    \subfigure[mr]{
        \includegraphics[width=.2\columnwidth]{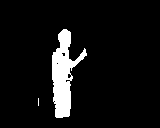}
        }    \hspace{-2.5mm}
    \subfigure[lb]{
        \includegraphics[width=.2\columnwidth]{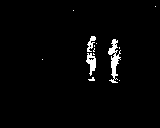}
        }    \hspace{-2.5mm}
    \subfigure[sc]{
        \includegraphics[width=.2\columnwidth]{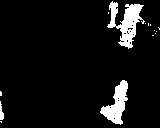}
        }    \hspace{-2.5mm}
    \subfigure[ap]{
        \includegraphics[width=.2\columnwidth]{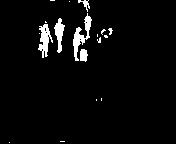}
        }    \hspace{-2.5mm}
     \subfigure[br]{
        \includegraphics[width=.2\columnwidth]{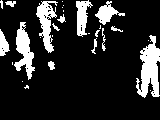}
        }    \hspace{-2.5mm}
    \subfigure[ss]{
        \includegraphics[width=.2\columnwidth]{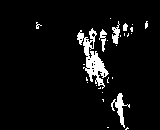}
        }  \hspace{-1.7mm}
}
	\caption{\small{Results on Li data set. From top to bottom: test images, the ground truth and the estimations of our model.}}
\label{fig_Li1}
\end{figure*}

\begin{table*}[t] \small
\caption{Results for Li, given as F-score.}
\label{tab:li2}
\begin{center}
\begin{tabular}{c|c|c|c|c|c|c|c|c|c|c}
\toprule
 methods & cam & ft& ws & mr & lb & sc  & ap & br & ss & mean\\
\midrule
Hist  & .1596& .0999& .0667& .1841& .1554& .5209& .1135& .3079& .1294 & .1930\\
MoG  & .0757& .6854& .7948& .7580& .6519& .5363&  .3335& .3838& .1388 & .4842\\
Map   & .6960& .6554& .8247& .8178& .6489 &.6677 & .5943 & .6019& .5770 & .6760\\
D-MoG   & .7624& .7265& .9134& .3871& .6665& .6721& .5663 &.6273 &.5269 & .6498\\
RPCA  & .5226& .8650 & .6082 & .9014 & .7245 & .7785 & .5879 & .8322 & .7374 & .7286\\
G-Lasso  & .8347 & .8789 & .9236 & .8995 & .6996 & .8019 & .5616 & .7475 & .6432 & .7767\\
\textbf{Ours}   &  \textbf{.8386} &  \textbf{.9011}& \textbf{.9424}$^*$& \textbf{ .9592}& \textbf{.8208}& \textbf{ .8500}& \textbf{ .7422}& \textbf{.8476}& \textbf{.7613} & \textbf{.8515}\\
\bottomrule
\end{tabular}
\end{center}
\end{table*}

\subsection{Comparison with State-of-the-Art}
\label{ssec:perf2}
A brief summary of all the models we compared can be found in Table \ref{tab:sum}. We compare our model to these models on all three data sets\footnote{Note that, since we are using the results reported by respective papers, not all the models have results on every sequence.}. Following the literature, for the ``Wallflower" data set, mis-classified number of pixels is used as the evaluation criteria; for both ``Li" and ``SABS," F-score (F) is used as the evaluation criterion. We put a ``$*$" on the upper-right corner of the scores to indicate that the sequence is of the SML case.

\noindent
\textbf{On Wallflower data set.} We tested our model on all the seven sequences of this data set.
In Table \ref{tab:wf2}, we provide quantitative comparisons, where our model achieved the least mis-classified number of pixels on five sequences and the second least on one sequence. Note, however, our model performed poorly on the sequence ``CF". The reason is that the foreground in ``CF" occupies a large portion of the tested frame, which violates the prior assumption on foreground sparsity. The same failure happened to both the RPCA and G-Lasso models, since both of them also assume sparse foreground prior. In Figure \ref{fig_wf1}, we show the qualitative results of our model on the seven sequences.

\noindent
\textbf{On Li data set.} We applied our model to all the nine sequences of the data set. In Table \ref{tab:li2}, we show quantitative comparisons, where our model achieved the highest F-score on all these sequences. Notably, in some sequences such as ``lb", ``ap" etc., the improvements over the second best are more than 10\%. On average, our model achieved an 8\% F-score gain ahead of the second best model. The qualitative results of all nine sequences are shown in Figure \ref{fig_Li1}.

\begin{figure}[h]
    \centering
    \subfigure[Test image]{
    \label{fig:demo1}
        \includegraphics[width=.3\columnwidth]{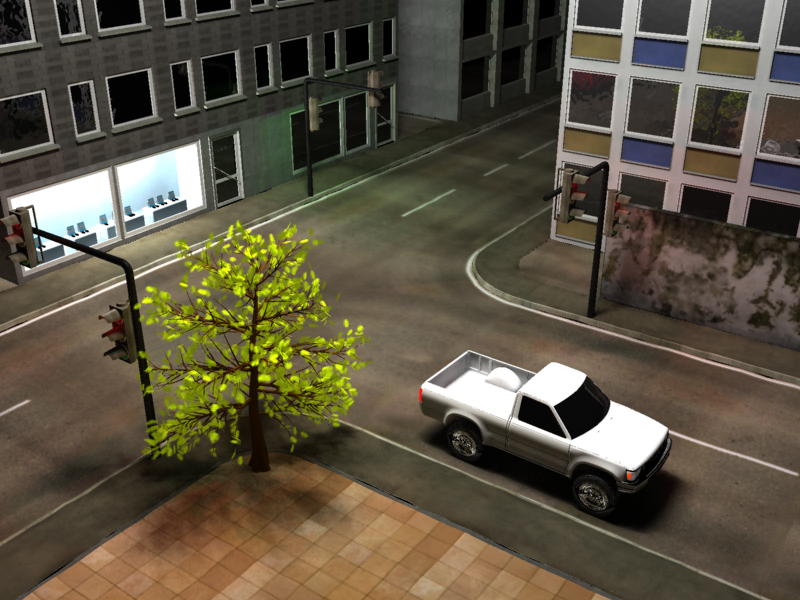}
        }    \hspace{-2.5mm}
    \subfigure[Ground Truth]{
        \includegraphics[width=.3\columnwidth]{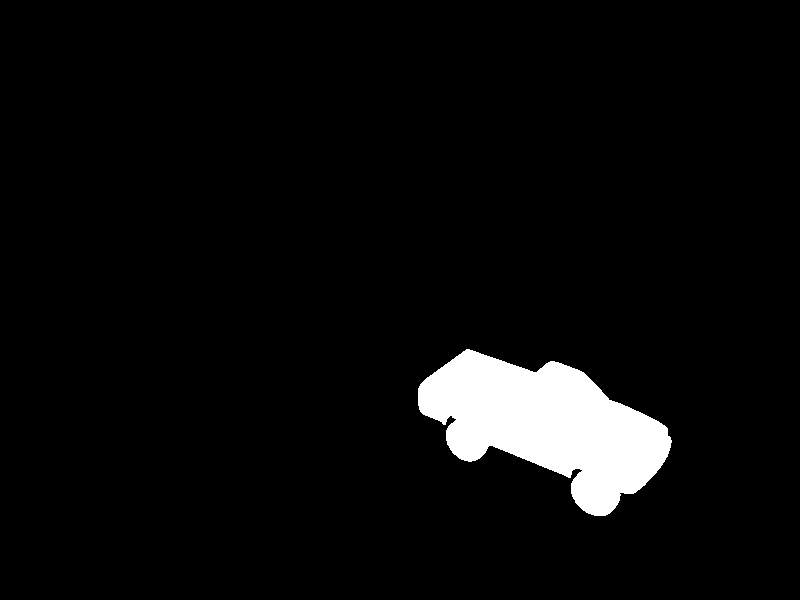}
        }   \hspace{-2.5mm}
    \subfigure[Ours: 0.876]{
        \includegraphics[width=.3\columnwidth]{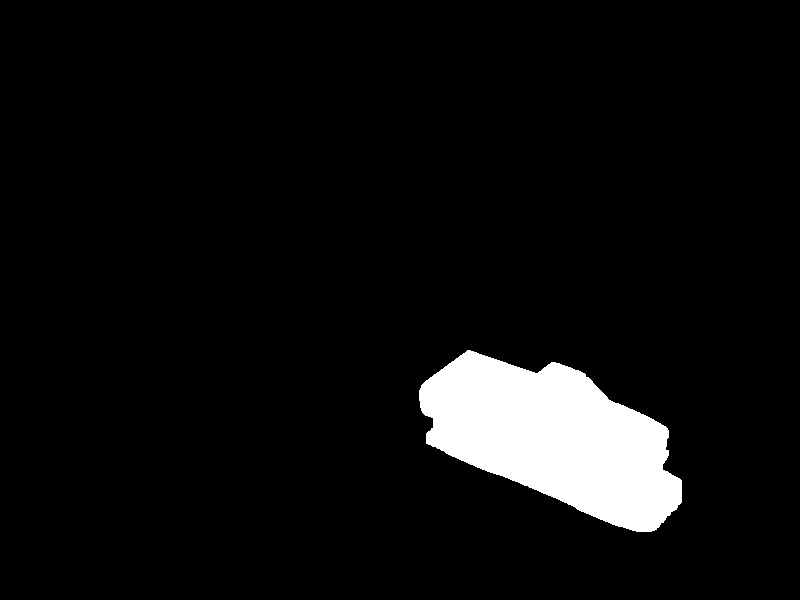}
        }
    \subfigure[G-Lasso: 0.848]{
        \includegraphics[width=.3\columnwidth]{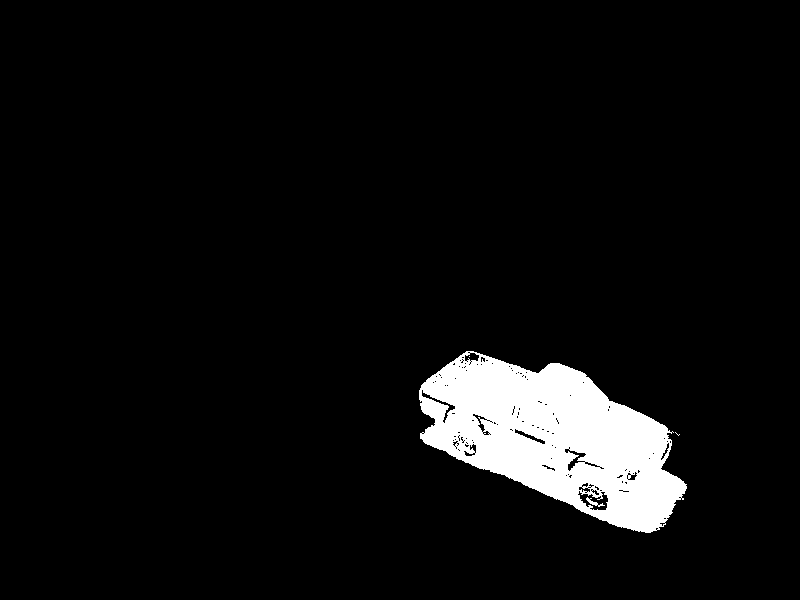}
        }  \hspace{-2.5mm}
    \subfigure[RPCA: 0.802]{
        \includegraphics[width=.3\columnwidth]{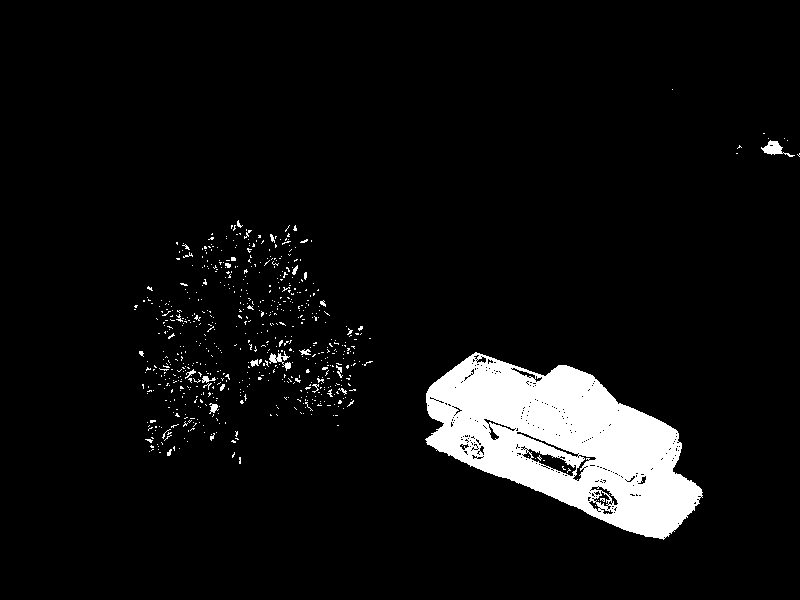}
        }  \hspace{-2.5mm}
    \subfigure[KDE: 0.803]{
    \label{fig:demo1}
        \includegraphics[width=.3\columnwidth]{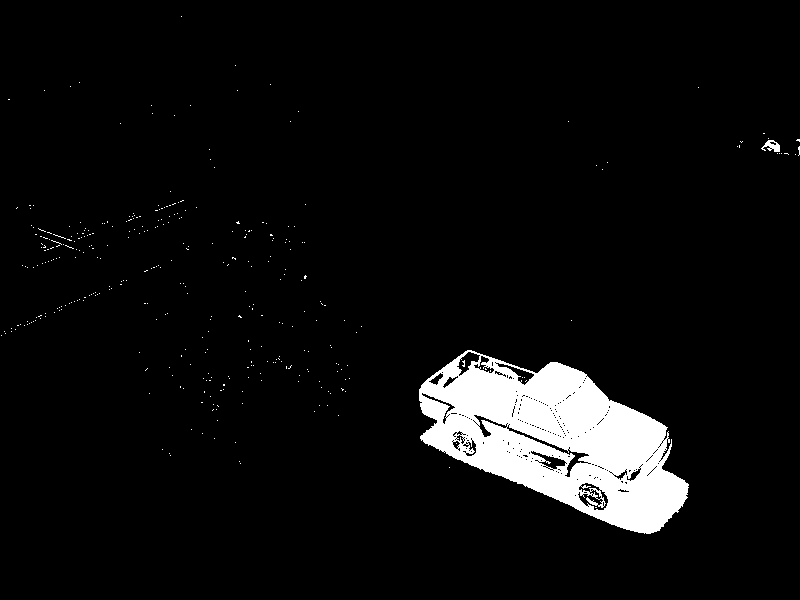}
        }
    \subfigure[Hist 0.782]{
        \includegraphics[width=.3\columnwidth]{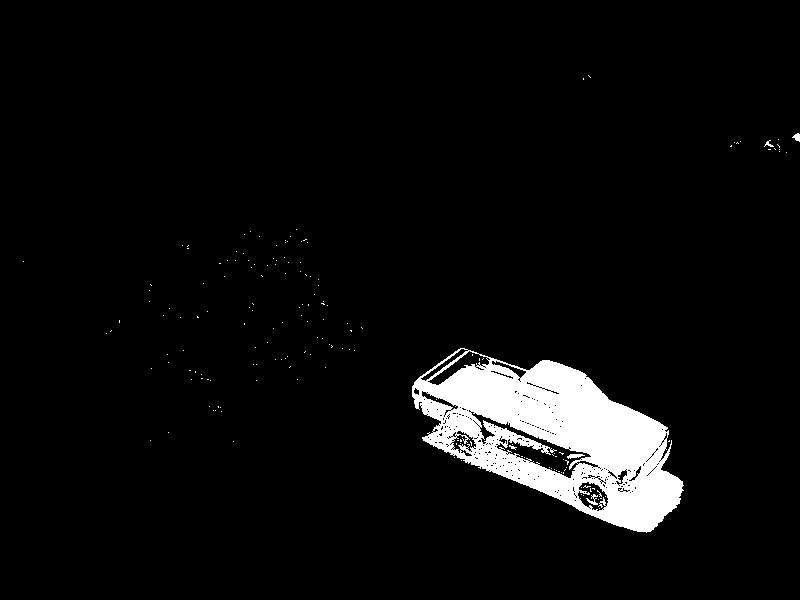}
        }    \hspace{-2.5mm}
    \subfigure[MoG: 0.819]{
        \includegraphics[width=.3\columnwidth]{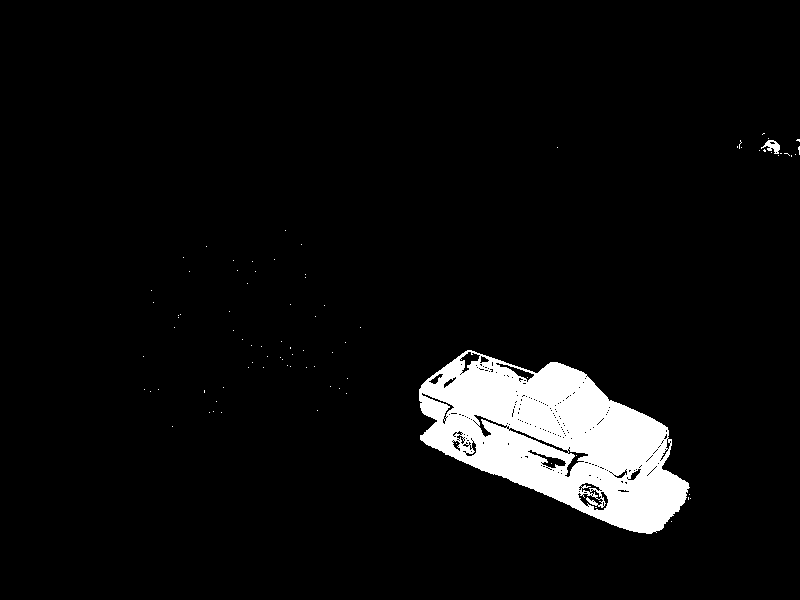}
        }   \hspace{-2.5mm}
    \subfigure[R-MoG: 0.807]{
    \label{fig:demo1}
        \includegraphics[width=.3\columnwidth]{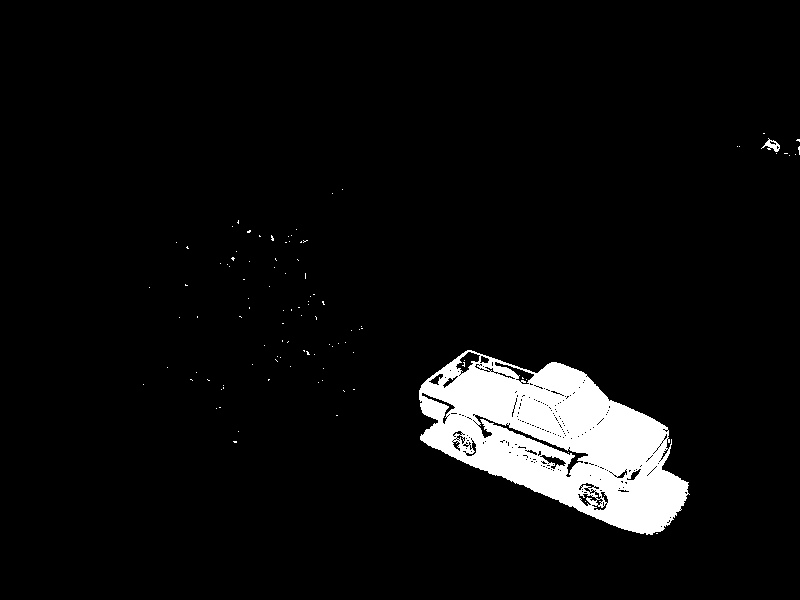}
        }
   \caption{\small{Results on the SABS data set. F-scores are shown.}}
\label{fig:bs_sabs}
\end{figure}

\noindent
\textbf{On SABS data set.} Following \cite{haines2012background}, we apply our model to the ``Basic" sequence and compared with the other models on this representative sequence. The results of different models on an example frame (No. 448) are illustrated in Figure \ref{fig:bs_sabs}. (The qualitative results on the whole sequence can be found on our webpage.) As is shown, our model almost cuts a perfect foreground (including its shadow). In the ground truth, the shadow is not included, which makes the value of the F-score relatively low. However, this definition of foreground may be controversial depending on the actually situations. Nevertheless, our model outperforms all the rest models on the test image. The average F-scores of all the models on the whole sequence are summarized in Table \ref{tab:accr1}, where our model is shown to have achieved the highest performance. 

\begin{figure}[h]
    \centering
    \subfigure[G-lasso ($3$$\times$$3$ block)]{
        \includegraphics[width=.3\columnwidth]{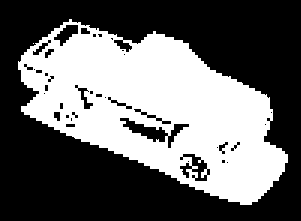}
        }
    \subfigure[G-lasso (coarse-to-fine superpixel)]{
        \includegraphics[width=.3\columnwidth]{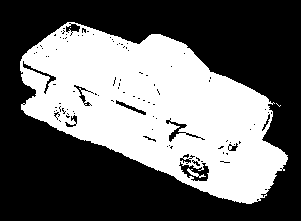}
        }
    \subfigure[Ours (adaptive fused lasso)]{
        \includegraphics[width=.3\columnwidth]{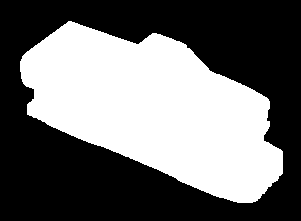}
        }
   \caption{\small{Different foreground regularization comparison. }}
\label{fig:bs_sabs2}
\end{figure}

\noindent
\textbf{Compare with group lasso.} As mentioned in the related works of Section \ref{sec:intro}, the group lasso regularization was applied to modeling foregrounds of BS in \cite{xu2013gosus} . The authors used both ``$3\times3$ blocks group" and ``coarse-to-fine superpixel group" structures to pursue connected sparse foregrounds. However, as can be seen from the above comparisons e.g. Table \ref{tab:li2}, \ref{tab:wf2} \& \ref{tab:accr1} and Figure \ref{fig:bs_sabs}, their performance are not as good as those of the proposed model. In Figure \ref{fig:bs_sabs2}, we provide a close-up comparison with the deliberately pre-defined grouping of pixels for foreground modeling. It shows that the group lasso model generates artifacts of detected foreground objects due to inappropriate pre-defined group structure. This arguably indicates that, compared to (adaptive) GFL, the group sparse models may not be flexible enough for recovering arbitrary foreground shapes.

\subsection{Discussions}
\label{ssec:discus}
\noindent
\textbf{Computation.} The algorithm does not take many iterations to converge, see e.g. Figure \ref{fig:alm}, and in practice the average number of iterations is about 10-20. Therefore, the major computational cost to pursue structured background and foregrounds in the mid-steps can be eased up by this few iterations.
Moreover, since the updating of the foreground are column-wise, the implementation can be highly paralleled in practice. 
The code can be downloaded on our webpage.

\noindent
\textbf{SML vs. UML.} Note that, in general when pure background frames are available, like most of the sequences in the Wallflower dataset, we have reported the results of the SML model. Such a choice outperforms its unsupervised counterpart, e.g. with an improvement of 24 (for WT) to 179 (for TD) pixels on the Wallflower dataset. However, this is not always the case. For example, in the ``cam" sequence of the Li dataset, although there are pure background frames, they seem to be less representative possibly due to some background changes. Then, the supervised model did not achieve obviously better results but still competitive, in this case: 0.8382 vs 0.8386. 

\noindent
\textbf{Comparison with Explicit Post-processing.} Arguably, explicit post-processing in BS e.g. \cite{nurhadiyatna2013background} can be viewed as a special case of foreground modeling since these methods are fundamentally using foreground structural priors to guide post-processing. Therefore, we carried out some experiments with the data used in \cite{nurhadiyatna2013background}, where MoG models are post-processed by a hole-filling method. In summery, our model achieved competitive or even better results, detailed comparisons can be found on our webpage.
. 

\section{Conclusion}
\label{sec:concl}

In this paper, we propose a method of background subtraction by exploiting structure information of the foregrounds to help background modeling. Our model works for both supervised and unsupervised learning paradigms and automatically pursue meaningful background and foregrounds. To optimize the new objective function, we proposed an effective algorithm by extending the ALM, which alternatively updates the background and the foreground matrices. Experimental results show that the proposed model achieves better than state-of-the-art performance on several popular public data sets.

\noindent
\textbf{Acknowledgements.} 
This work was supported in part by 973 grant 2015CB351800 and NSFC grants 61272027, 61231010, 61421062, 61210005.

{\small
\bibliographystyle{ieee}
\bibliography{ref}
}

\end{document}